\documentclass[preprint,12pt,3p]{elsarticle}

\usepackage{amssymb}
\usepackage{amsmath}
\usepackage{mathtools}
\usepackage{array}
\usepackage{cancel}
\usepackage{bm}
\usepackage{bbm}
\usepackage[usenames, dvipsnames]{color}
\usepackage{algorithm,algorithmic}
\usepackage[toc,page]{appendix}
\usepackage{caption}
\usepackage{subcaption}
\usepackage{hyperref}
\usepackage{multicol}
\usepackage{textcomp}
\usepackage[hang,flushmargin]{footmisc}
\usepackage{siunitx}
\usepackage{mdframed}
\usepackage{listings}
\usepackage{sidecap}
\usepackage{soul}
\usepackage{xcolor}

\DeclareMathOperator*{\argmax}{arg\,max}

\makeatletter
\providecommand{\doi}[1]{%
  \begingroup
    \let\bibinfo\@secondoftwo
    \urlstyle{rm}%
    \href{http://dx.doi.org/#1}{%
      doi:\discretionary{}{}{}%
      \nolinkurl{#1}%
    }%
  \endgroup
}
\makeatother

\newcommand{\BlackBox}{\rule{1.5ex}{1.5ex}}  



\usepackage{graphicx}
\graphicspath{{./figures/}}
\DeclareGraphicsExtensions{.pdf,.jpeg,.png}

\journal{Information Sciences}

\begin{document}

\begin{frontmatter}

\title{On the Effect of Suboptimal Estimation of Mutual Information in Feature Selection and Classification}

\author[label1]{Kiran Karra\corref{cor1}}
\address[label1]{Virginia Tech \\ Department of Electrical and Computer Engineering \\ 900 N. Glebe Rd., Arlington, VA, 22203}

\cortext[cor1]{Corresponding Author}
\ead{kiran.karra@gmail.com}
\ead[url]{kirankarra.wordpress.com}

\author[label1]{Lamine Mili}
\ead{lmili@vt.edu}

\begin{abstract}
This paper introduces a new property of estimators of the strength of statistical association, which characterizes how well an estimator rank orders dependencies between continuous and discrete random variables.  The new property, termed the estimator response curve (ERC), is easily computable and provides a marginal distribution agnostic way to assess an estimator's performance.  In the context of rank ordering dependencies, it overcomes drawbacks of current metrics of assessment, including statistical power, bias, and consistency.  We utilize the ERC to compare various measures of the strength of association that satisfy the data processing inequality (DPI), and show that the \textit{CIM's} ERC compares favorably to \textit{kNN}, \textit{vME}, \textit{AP}, and $H_{MI}$ estimators of mutual information.  This rank ordering of estimator performance is then shown to be positively correlated to performance in feature selection and classification tasks on various real world datasets, indicating that the ERC is a valuable tool in assessing how an estimator of mutual information may affect end-to-end machine learning performance.
\end{abstract}


\end{frontmatter}

\section{Introduction}
Many applications of data mining and machine learning utilize measures of the strength of association between random variables to reduce data redundancy and find interesting associations within datasets.  Measures of statistical association, such as the correlation coefficient $\rho$ \cite{corrcoef}, \textit{MIC} \cite{mic}, the \textit{RDC} \cite{rdc}, the \textit{dCor} \cite{dcorr}, the \textit{Ccor} \cite{ccorr_cite}, \textit{CoS} \cite{cos_cite}, and \textit{CIM} \cite{cim} and estimators of mutual information such as the \textit{kNN} \cite{knn_mi}, the \textit{vME} \cite{vonMisesMI}, and the \textit{AP} \cite{shannonapMI} are used to compute these associations.  Although the theoretical forms of these measures of association are generally applicable, the estimators of these quantities often fall short when faced with real world impediments.  Characteristics such as whether the data are discrete or continuous, linear or nonlinear, skewed or balanced, monotonic or nonmonotonic, noisy or clean, and independent and identically distributed (\textit{i.i.d.}) or serially dependent, to name a few, drive the performance of the estimator.  Because the estimation of the strength of association is often abstracted from the algorithms which rely on them, more emphasis in machine learning research is currently placed on designing and developing new algorithms rather than more accurate estimation.  

In this paper, we develop a new property of estimators of the strength of statistical association, which characterizes how well an estimator rank orders dependencies between continuous and discrete random variables.  The new property, termed the estimator response curve (ERC), is easily computable and provides a marginal distribution agnostic way to assess an estimator’s performance.  We then utlize the ERC to show the effect of suboptimal estimation of mutual information on feature selection and classification performance.  We focus on the scenario where the strength of association needs to be measured between noisy \textit{i.i.d} continuous and discrete random variables (henceforth referred to as hybrid random variables) that are skewed, where the number of unique outcomes of the discrete random variable are small and the dependence structures are nonlinear.  This case represents an important subset of problems in machine learning, where real world datasets that often have nonlinear associations between them with skewed marginal distributions, need to be classified according to provided output labels.  Examples abound in many domains, including medicine where data is used to predict the presence or absence of a disease, or finance where transactional data are used to predict whether a particular purchase is fraudulent or not, to name a few.

In this manuscript, we restrict ourselves to only compare estimators that are proven to satisfy the data processing inequality (DPI); that is, all estimators of mutual information (\textit{kNN},\textit{vME},\textit{AP},$H_{MI}$) and the index \textit{CIM}.  DPI satisfying measures are preferred in machine learning due to the relationship between the DPI and Markov chains \cite{mic_not_equitable}.  Furthermore, the DPI assumption is implicit in many machine learning algorithms that utilize measures of the strength of association, such as the maximum-relevance minimum-redundancy (MRMR) algorithm for Markov network discovery and feature selection \cite{aracne, mrmr}.

The paper is organized as follows.  We begin with an introduction to estimation of the strength of association between random variables and stochastic processes. Here, we discuss the difficulty in estimating the strength of association for the hybrid random variable scenario.  In addition, we formally define the ERC.  Then, we perform synthetic data simulations that measure the ERC of the various estimators under consideration.  To accomplish this, we utilize the copula framework to generate data with a wide range of dependence structures and marginal distributions to comprehensively test the performance of these estimators under various real world impediments.  We show that for the properties tested, the \textit{CIM} estimator's ERC compares favorably to other DPI satisfying measures of mutual information.  Next, we apply these various estimators to real-world datasets and characterize how suboptimal estimation of mutual information affects machine learning classification performance.  We synthetically vary these real world datasets further in order to simulate more extreme scenarios that may be encountered, and characterize the performance through these variations.  The \textit{CIM} estimator's performance is shown to compare favorably to other DPI satisfying measures of association when used for these real-world datasets.  This corroborates our findings with synthetic data, which empirically indicates the usefulness of the estimator response curve.  Concluding remarks are provided in the final section.

Random variables $X$ and $Y$ are said to be dependent if $f_{XY}(x,y) \neq f_X(x)f_Y(y)$, where $f_{XY}$ is the joint distribution of $X$ and $Y$, and $f_X$ and $f_Y$ are the marginal distributions of $X$ and $Y$, respectively.  The strength of that dependence can be viewed from two aspects.  From an information theory perspective, the strength of the dependence encompassed by the joint density, $f_{XY}$, can loosely be stated as inversely proportional to the amount of disorder in the joint density.  From a statistical perspective, the strength of the dependence can be viewed as a generalized distance measure between statistical independence, where $f_{XY} = f_X \times f_Y$, and the true joint distribution $f_{XY}$.

The first solution put forth to measure this strength of dependence was the correlation coefficient, $\rho$.  Although popular, the correlation coefficient has many drawbacks, with the most notable being that it can only measure linear dependence.  Many real world datasets have nonlinear dependence structures, and the correlation coefficient does not fully capture the strength of association between the random variables in these scenarios.  Measures such as distance correlation, \textit{dCor} \cite{dcorr}, and the maximal information coefficient, \textit{MIC} \cite{mic}, were then introduced by researchers in order to overcome the linearity limitation of the correlation coefficient.  

Another class of estimators of the strength of association between random variables and stochastic processes utilize the copula framework to overcome the linearity limitation.  Copulas are multivariate joint probability distribution functions for which the marginal distributions are uniform \cite{nelsen, Embrechts01}.  The existence of a copula $C(\cdot)$ associated with a collection of random variables, $X_1, \dots, X_n$, following a joint probability distribution function, $F(\cdot)$, and marginals $F_{X_i}(x_i)$ is ensured by Sklar's theorem, which states that

\begin{equation} \label{eq:sklar1}
F(x_1, \dots, x_n) = C(F_{X_1}(x_1), \dots, F_{X_n}(x_n)).
\end{equation}
This theorem guarantees the unicity of the copula $C$ for continuous random variables and it unveils its major property, which is its ability to capture the unique dependency structure between any continuous marginal distributions $X_1, \dots, X_n$.  Thus, the copula $C$ can be used to define a measure of dependence between continuous random variables \cite{scarsini}.  Nonmonotonic measures of dependence based on the copula framework include Kendall's $\tau$ \cite{kendalltau}, variants of $\tau$ to account for ties in the data such as $\tau_b$ and $\tau_{vl}$ \cite{taub, tauvl}, and Spearman's $\rho$ \cite{spearmansrho}. However, because copulas capture the full dependence structure between random variables, they can also be designed to capture non-monotonic dependence.  This concept has been utilized by various indices, including copula correlation coefficient, \textit{Ccor} \cite{ccorr_cite}, the copula statistic, \textit{CoS} \cite{cos_cite}, the randomized dependence coefficient, \textit{RDC} \cite{rdc}, and the copula index for monotonicity \textit{CIM} \cite{cim}.  

The advantage of these aforementioned measures of association is that they are true indices; that is, that they take on values between zero and one, where one represents perfect association between the random variables or stochastic processes, and zero represents no association.  For estimators that also satisfy R\'enyi's properties of dependence measures, a value of zero also implies statistical independence \cite{Renyi1959}.  The notable disadvantage of all of these indices is that all of them, except for the \textit{CIM}, are not necessarily proven to satisfy certain desirable properties of estimators of association.  These properties include R\'enyi's properties \cite{Renyi1959}, and the data processing inequality (DPI) \cite{mic_not_equitable}.  For this reason, machine learning practitioners often use mutual information (MI) as a measure of the strength of association between two random variables $X$ and $Y$, which is defined as 
\begin{equation}\label{eq:mi}
I(X,Y) = \int_Y \int_X f_{XY}(x,y) \mathbf{log} \frac{f_{XY}(x,y)}{f_X(x)f_Y(y)} dx dy,
\end{equation}
where $f_{XY}(x,y)$ is the joint distribution of $X$ and $Y$, and $f_X(x)$ and $f_Y(y)$ are the marginal distributions of $X$ and $Y$, respectively \cite{CoverInfoTheo}.  Four common estimators of mutual information include k-nearest neighbors, \textit{kNN} \cite{knn_mi}, von Mises Expansion, \textit{vME} \cite{vonMisesMI}, adaptive partitioning \textit{AP} \cite{shannonapMI}, and entropy based estimation, hereby denoted in this paper as $H_{MI}$.  Briefly, \textit{kNN}-based estimation of mutual information uses the k-nearest neighbors approach to estimate the univariate and multivarite densities in (\ref{eq:mi}), and then applies the integral or summation in the discrete case.  The adaptive partitioning approach to estimating mutual information uses an algorithm to optimally partition the space spanned by the two random variables, such that mutual information can be accurately estimated \cite{shannonapMI}.  Both the \textit{kNN} and \textit{AP} approaches attempt to accurately measure mutual information through partitioning of the space.  Conversely, von Mises expansion utilizes influence functions to measure the mutual information between random variables.  Finally, the entropy based estimator $H_{MI}$~\cite{micite} uses the relationship between mutual information and conditional entropy, given by

\begin{equation}\label{eq:h_mi}
I(X,Y) = H(Y) - H(Y|X),
\end{equation}
where
\begin{equation}\label{eq:h}
H(X) = - \int_{-\infty}^{\infty} p(x) \mathbf{log}[p(x)] dx
\end{equation}

\subsection{Where do these estimators fall short?}
When measuring the strength of association between continuous and discrete random variables, most of the estimators previously mentioned fall short.  In general, it becomes more difficult to measure association between continuous and discrete random variables as the number of unique discrete outcomes decreases \cite{genest}.  The case of measuring the strength of association between hybrid random variables, however, is extremely important in machine learning.  From classification, clustering, and feature selection perspectives, features are typically amenable to being modeled as continuous random variables, while outcomes or clusters are better modeled as discrete random variables.

\begin{figure}[!t]
\centering
\includegraphics[width=2.5in]{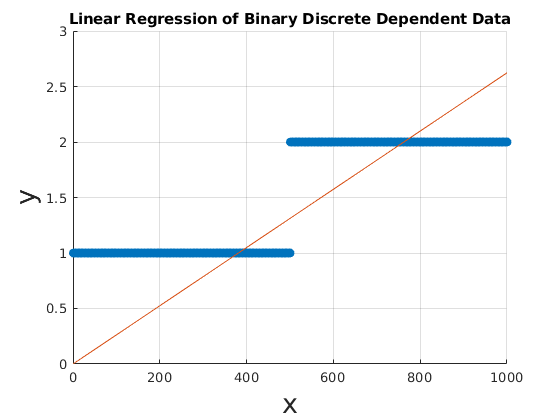}
\caption{Linear Regression of continuous independent variable, $X$, and discrete dependent variable $Y$ with only two unique outcomes.  Here, $X$ and $Y$ are perfectly associated, but the correlation coefficient is computed to be $0.86$.}
\label{fig:ols_bddd}
\end{figure}

Each of the aforementioned estimator classes suffers from difficulties in dealing with hybrid random variable scenarios.  The correlation coefficient, $\rho$, is actually the standardized linear regression coefficient between random variables $X$ and $Y$.  If $Y$ takes on a small number of unique outcomes, the MMSE objective for solving the regression coefficient does not properly capture the dynamics of the data, and in fact violates an implicit assumption of linear regression, that the dependent variable, $Y$ be continuous.  This is illustrated in Fig.~\ref{fig:ols_bddd}.  In it, the independent random variable, $X$, and the dependent random variable, $Y$ are perfectly associated; the rule 
\begin{equation*}
    Y = 
    \begin{cases}
        0, & \text{for } x\leq 500\\
        1, & \text{for } x> 500
    \end{cases}
\end{equation*}
describes the functional relationship in Fig.~\ref{fig:ols_bddd}.  However, the correlation coefficient is measured to be $0.86$.  

As for the Maximal Information Coefficient, \textit{MIC} and other mutual-information-based methods such as \textit{kNN}, \textit{AP}, and $H_{MI}$, discretization of the continuous random variable is required in order to apply formulas such as (\ref{eq:mi}) and (\ref{eq:h_mi}).  However, this technique cannot be performed optimally without taking into account the end goals of the discretized data, and in addition, it results in information loss \cite{discretizeieee}.

The copula-based estimators mentioned above are also not immune to the hybrid scenario.  When modeling discrete random variables with copulas, Sklar's theorem  does not guarantee the unicity of the copula $C$ and many copulas satisfy (\ref{eq:sklar1}) due to ties in the data~\cite{neslehova, genest, denuit}.  This ambiguity is the reason why copula-based estimators also experience some difficulty in measuring association between continuous and discrete random variables.  The exception to this is \textit{CIM}, which is based on a bias-corrected measure of concordance, $\tau_{KL}$ that can account for continuous and discrete data simultaneously~\cite{cim}.

\subsection{The Effect of Suboptimal Estimation}
All aforementioned estimators make errors when estimating the strength of association between hybrid random variables.  These errors manifest as incorrect rank ordering of features, which can affect feature selection and classification performance in machine learning pipelines.   In order to help predict how misestimation will affect rank ordering of dependencies, we introduce the estimator response curve (ERC).  The ERC is defined as the relationship between the actual strength of association between random variables, $X$ and $Y$, and the estimated strength of association between $X$ and $Y$, over the entire range of possible strengths of statistical association. It's motivation, properties, and development are detailed in the next section.

\section{Estimator Response Curve (ERC)}
In this section, we develop and describe the estimator response curve (ERC).  We begin by describing some previously developed metrics of estimator performance, and show why they do not fully capture the performance of an estimator.  We then develop and characterize the estimator response curve.

There are several properties of estimators of the strength of association which are important in characterizing its performance.  Theoretically provable properties include R\'enyi's seven properties of a dependence measure and the data processing inequality (DPI) for dependence measures~\cite{Renyi1959, mic_not_equitable}.  R\'enyi's seven properties of a measure of dependence are important because they establish the space over which the estimator can be used, the exchangeability of the random variables under consideration, and the ranges of possible outputs of the estimators.  Similarly, DPI is important for an estimator because it ensures that, when measuring the strength of association between random variables in a causal chain, indirect causes measure weaker than more direct causes~\cite{mic_not_equitable}.  It is interesting to note that most estimators of mutual information, such as \textit{kNN}, \textit{vME}, \textit{AP}, and $H_{MI}$ do not satisfy R\'enyi's seven properties.  Conversely, most indices of dependence, including \textit{dCor}, \textit{Ccor}, \textit{CoS}, and \textit{RDC} are not proven to satisfy the DPI.  The notable exception here is the \textit{CIM}, which is proven to satisfy the DPI under certain conditions~\cite{cim}.

Important empirical properties of an estimator include the statistical power, bias, and consistency.  Statistical power is the likelihood that the estimator measures an association between the random variables, when the random variables are statistically dependent.  Usually, the power of an estimator is characterized across a range of dependencies and noise levels to fully assess the estimator's ability to detect different types of dependencies \cite{cim}.  The importance of power, especially under linear dependence, was originally outlined in Simon and Tibshirani's work \cite{simonandtibs}.  Statistical bias is the difference between the true value and the estimated value of the quantity to be measured, and can be defined mathematically as $\text{Bias}_{\theta}[\hat{\theta}] = E_{x|\theta}[\hat{\theta}] - \theta$, where $\theta$ is the true value of the quantity to be estimated, $\hat{\theta}$ is the estimated value, and $E_{x|\theta}$ is the expected value over the conditional probability distribution $P(x|\theta)$.  However, bias is typically only computed under the scenario of independence, where it is known that the value of $\theta$ should be 0.  Finally, statistical consistency measures the asymptotic properties of the estimator; an estimator is said to be consistent if the estimated value approaches the true value of the estimator as the sample size grows.  Stated mathematically, an estimator of $T_n$ of $\theta$ is said to be consistent if $\underset{n\rightarrow \infty}{\text{plim}} T_n = \theta$.

While these three empirical properties are important to assess an estimator's performance, none of them capture the concept of the rate of change of an estimated value, as the strength of dependence between the random variables changes.  If the rate of increase (or decrease) of an estimated value is not proportional to the rate of increase (or decrease) in the dependence strength between the random variables, there is a nonzero likelihood that the estimator will incorrectly rank the strength of associations between features and output classes in supervised learning.  This becomes especially important in the hybrid random variable scenario, where it is already more difficult to measure the strength of association between two random variables for the reasons discussed previously~\cite{genest}.  These rates of increase determine the estimators ability to distinguish stronger from weaker relationships, when both relationships are statistically significant.  We term the relationship between the actual strength of association between the random variables, $X$ and $Y$, and the estimated strength of association between $X$ and $Y$ over the entire range of possible strengths of statistical association to be the response curve of an estimator.  This curve can help explain how an estimator will perform when multiple strengths' of associations need to be measured and ranked, as in mutual information based Markov network discovery and feature selection \cite{aracne, mrmr, mrnet}.  

An ideal estimator would increase (or decrease) its estimate of the strength of association between random variables $X$ and $Y$ by $\hat{\Delta}$, due to a corresponding increase (or decrease) of $\Delta$ of the strength of association between $X$ and $Y$, across the full range of possible dependence between random variables.  The ratio of $\hat{\Delta}$ to $\Delta$ determines the shape of the ERC, and predicts how a particular estimator will perform when it's outputs are used to rank order dependencies between random variables. If it is desirable to more accurately distinguish stronger dependencies than weaker ones, the ideal response of an estimator across the full range of possible dependencies is a monotonically increasing convex function, with the degree of convexity directly proportional to an increased ability of the estimator to distinguish stronger dependencies apart.  This scenario corresponds to $\hat{\Delta} > \Delta$ when the strength of association is high.  Conversely, if it is desirable to more accurately distinguish weaker dependencies than stronger ones, the ideal response of an estimator across the full range of possible dependencies is a monotonically increasing concave function, with the degree of concavity directly proportional to an increased ability of the estimator to distinguish weaker dependencies apart.  This scenario corresponds to $\hat{\Delta} < \Delta$ when the strength of association is high.  The special case of $\hat{\Delta} = \Delta$ is ideal, where the estimator is able to distinguish all dependence types equally well.  However, even if an estimator has this kind of response curve, its variance must be low to have a high likelihood that dependencies will be correctly ranked.  

\begin{figure}[!t]
\centering
\includegraphics[width=2.5in]{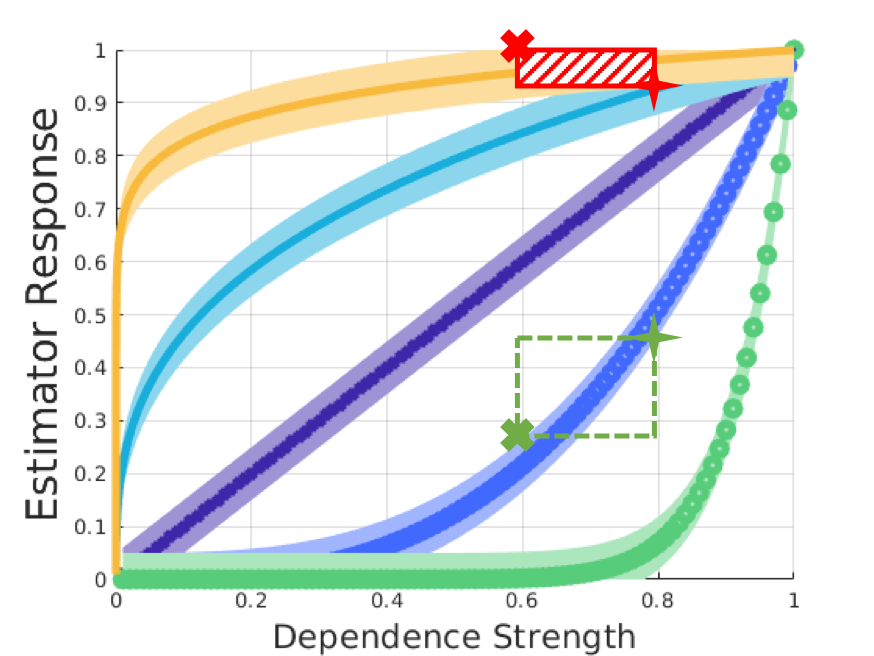}
\caption{Illustration of the various estimator response curves.  The linear response is shown in purple; in it, the estimator attempts to distinguish between all strengths of dependence equally, while in the convex curves shown with $o$ markings in green and blue, stronger dependencies are have a higher likelihood of being ranked correctly.  Conversely, in the concave response curves denoted with marks in teal and yellow, the estimator has a higher likelihood of ranking weaker dependencies correctly.  The red shaded rectangle shows a region of non-zero probability that an estimator with the yellow response curve would have in misranking dependence strengths between two pairs of random variables having strengths of association to be $0.6$ and $0.8$, respectively.  The green hollow rectangle shows the region of zero probability that an estimator with the blue response curve would have in misranking dependence strengths between two pairs of random variables having strengths of association to be $0.6$ and $0.8$, respectively.}
\label{fig:response_curve}
\end{figure}

Various hypothetical response curves are shown in Fig.~\ref{fig:response_curve}.  The linear response is shown in purple; in it, the estimator attempts to distinguish between all strengths of dependence equally, while in the convex curves shown with $o$ markings in green and blue, stronger dependencies are have a higher likelihood of being ranked correctly.  Conversely, in the concave response curves denoted with marks in teal and yellow, the estimator has a higher likelihood of ranking weaker dependencies correctly.  The curve is scale-invariant, because it examines the rates of change of an estimator, rather than absolute values.  It also shows that nonlinear rescaling of an estimators output may affect its ability to correctly rank strengths of association.  An example of this is Linfoot's informational coefficient of correlation \cite{linfoot}, which attempts to rescale mutual information to be in the range of $0$ to $1$, in order to be used as a bounded measure of the strength of association between random variables $X$ and $Y$. The rescaling function is given by

\begin{equation*}
    r(X,Y) = \sqrt{1-e^{-2 I(X,Y)}},
\end{equation*}
where $I(X,Y)$ is the mutual information.  Depending on the variance of the estimator (explained in further detail below), this nonlinear scaling could have an adverse affect on ranking the strengths of association.

The curves in Fig.~\ref{fig:response_curve} also show the variance of the estimated quantity as a function of the strength of dependence.  The variance, along with the concavity/convexity of the estimator determines the probability of correctly ranking dependencies between different pairs of random variables.  More specifically, the probability of correctly ranking two different pairs of random variables according to their strength of association is inversely proportional to the area encompassed by the rectangle covering the space between the maximum possible value the estimator can take on for the weaker dependency, denoted by $\hat{\theta}^{\text{max}}_{\text{weaker}}$, and the minimum possible value the estimator can take on for the stronger dependency, denoted by $\hat{\theta}^{\text{min}}_{\text{stronger}}$, if $\hat{\theta}^{\text{max}}_{\text{weaker}} > \hat{\theta}^{\text{min}}_{\text{stronger}}$.  For example, in Fig.~\ref{fig:response_curve}, suppose that the true value of the strength of association between $X_1$ and $Y$ is $0.6$, and the true value of the strength of association between $X_2$ and $Y$ is $0.8$, and our goal is to rank them according to their strengths, using an estimator.  If the estimator had a response curve similar to the blue curve, then the probability of misranking these dependencies is zero here, because $\hat{\theta}^{\text{max}}_{\text{weaker}}$, denoted by the green $\times$ symbol is less than $\hat{\theta}^{\text{min}}_{\text{stronger}}$, denoted by the green four pointed star.  Conversely, if the estimator had a response curve similar to the yellow curve, then $\hat{\theta}^{\text{max}}_{\text{weaker}}$ denoted by the red $\times$ symbol is greater than $\hat{\theta}^{\text{min}}_{\text{stronger}}$, denoted by the red four pointed star.  The probability of misranking these dependencies is nonzero and is proportional to the area given by the area in the red shaded rectangle in Fig.~\ref{fig:response_curve}.  These probabilities do not need to be exactly computed, but identifying them helps to understand how an estimator may perform for these applications.  In summary, the estimator response curve is a quick, marginal distribution invariant method to assess how the estimator will perform under various dependencies.  

\subsection{Guidelines for Computing the ERC}
The ERC is a useful tool for determining the efficacy of a certain estimator for measuring the strength of statistical association in a range of applications.  However, under the definition provided above, one needs to have the true dependence strength to construct the ERC, which is not usually available for real world data.  In these scenarios, we take the approach of constructing a surrogate ERC. Two types of surrogate ERCs can be constructed: 1) tailored towards a specific dataset and can best predict performance on a specific dataset of interest, and 2) a general ERC that can generally assist in predicting how an estimator may perform on a wide variety of datasets, while lacking specific information about any specific dataset.

To construct a surrogate ERC tailored towards a specific dataset, one needs to first identify the monotonicity structure of the dataset to be analyzed and processed.  The monotonicity structure of dependencies refers to the number of monotonic regions of association between two random variables. The dependence structures are then mapped to a copula-based simulation framework that can accurately assess how the various estimators under consideration will perform for the dataset under consideration.  After creating the synthetic data, the ERC can be computed and the optimal estimator selected.  Listing 1 below details these steps.  This methodology generally follows the recommendations of \cite{madsenbirkes} when realizations of joint distributions with nonlinear dependence relationships.

\begin{mdframed}
\textbf{Listing 1: Guidelines on computing ERC tailored to a dataset}
\begin{enumerate}
    \item Determine monotonicity structure of pair-wise dependencies of dataset using algorithms such as $CIM$~\cite{cim}. \footnote{If this is not feasible, assume monotonic dependence structures.  This assumption is justified by a large scale study conducted, where it was found that across three different areas of science, and comparing over $75000$ dependence structures, atleast $96\%$ of them were found to be monotonic.~\cite{cim}}
    \item Compute Pearson's skewness measure for all continuous marginal distributions, and the probability mass functions for discrete distributions.
    \item Simulate random numbers from copulas which match the monotonicity structure in the datasets to be analyzed, over the range of dependence afforded by the copula family. \footnote{For monotonic dependence, the Gaussian, T, Frank, Gumbel, and Clayton copulas capture a wide range of dependence structures.}
    \item Utilizing Sklar's Theorem (\ref{eq:sklar1}), generate random variates $X$ and $Y$ which have the prescribed dependence structure over the range of strengths of association captured by the copula, and marginal distributions as captured by Step 2.
    \item Compute estimator responses for each dataset over the range of estimators considered for usage in the application, and plot the response of each estimator against the strength of dependence captured by the associated synthetically generated dataset.
    \item Identify the estimator which has ideal response curve, based on the criterion of which subsets of dependence strengths' need to be most accurately rank ordered. 
\end{enumerate}
\end{mdframed}

If a general ERC is to be constructed, the ERC for the Gaussian, T, Frank, Gumbel, and Clayton copulas using skewed marginal distributions should be simulated.  This would yield a worst case scenario of the ERC performance for the estimators tested under a wide variety of monotonic dependence structures.

\section{Synthetic Simulations}\label{sec:synthetic_simulations}
In this section, we detail simulations conducted to compute the estimator response curve of the aforementioned estimators of the strength of association that satisfy the DPI constraint.  To accomplish this, we use the copula framework, which allows us to generate various linear and non-linear dependency structures between random variables, while being agnostic to the marginal distributions.  We take the recommendations of Listing 1, and compute the ERC using the five copula types for positively skewed, no-skewed, and negatively skewed marginal distributions, with the dependent variable being discrete and having the support $\{0, 1\}$ which corresponds to binary classification. 

Our synthetic simulation framework utilizes the Gaussian, T, Frank, Gumbel, and Clayton copulas.  These copulas represent different kinds of dependence structures, with the Gaussian modeling linear dependence, and the remaining copulas modeling non-linear dependence patterns such as tail dependence.  Each of the copulas has a single parameter, $\theta$, which controls the strength of dependence, and corresponds directly to the mutual information contained within the dependence structure.  Because the scale and support of $\theta$ varies between different copula families, our simulations modulate the strength of dependence between the random variables through the rank correlation measure, $\tau$, which has a one-to-one correspondence to $\theta$ for every copula that was simulated.  After picking a certain copula family, we generate random variates $u$ and $v$ for all values of $\theta$.  Next, we apply the inverse transform $F^{-1}(U) = X$ and $G^{-1}(V) = Y$ respectively to generate $x$ and $y$.  

To simulate the machine learning scenario of continuous explanatory features and discrete outcomes, we choose $X$ to be a continuous random variable and $Y$ to be a discrete random variable.  The three scenarios considered are when $X$ and $Y$ are both skewed positively, not skewed, and both skewed negatively.  In the positive skew scenario, we choose $X$ to be a Beta distribution with $\alpha=12, \beta=2$.  In the negative skew situation, we choose $X$ to be a Beta distribution with $\alpha=2, \beta=12$.  In the no skew situation, we choose $X$ to be a Beta distribution $\alpha=10, \beta=10$.  Similarly, in the left skew situation, we choose $Y$ to be a discrete distribution, with a probability mass function taking on the vector $[0.72,0.28]$.  In the right skew situation, the probability mass function of $Y$ is given by the vector $[0.28,0.72]$.  Finally, in the no skew situation, the probability mass function of $Y$ is given by $[0.5,0.5]$.  In all these scenarios, the cardinality of the support set of $Y$ is two.  This corresponds to the binary classification scenario.  Additionally, all skewed distributions described above correspond to a skewness magnitude of $1$.

\begin{figure*}
    \centering
    \begin{subfigure}[t]{1.\textwidth}
        \centering
        \includegraphics[height=1.4in]{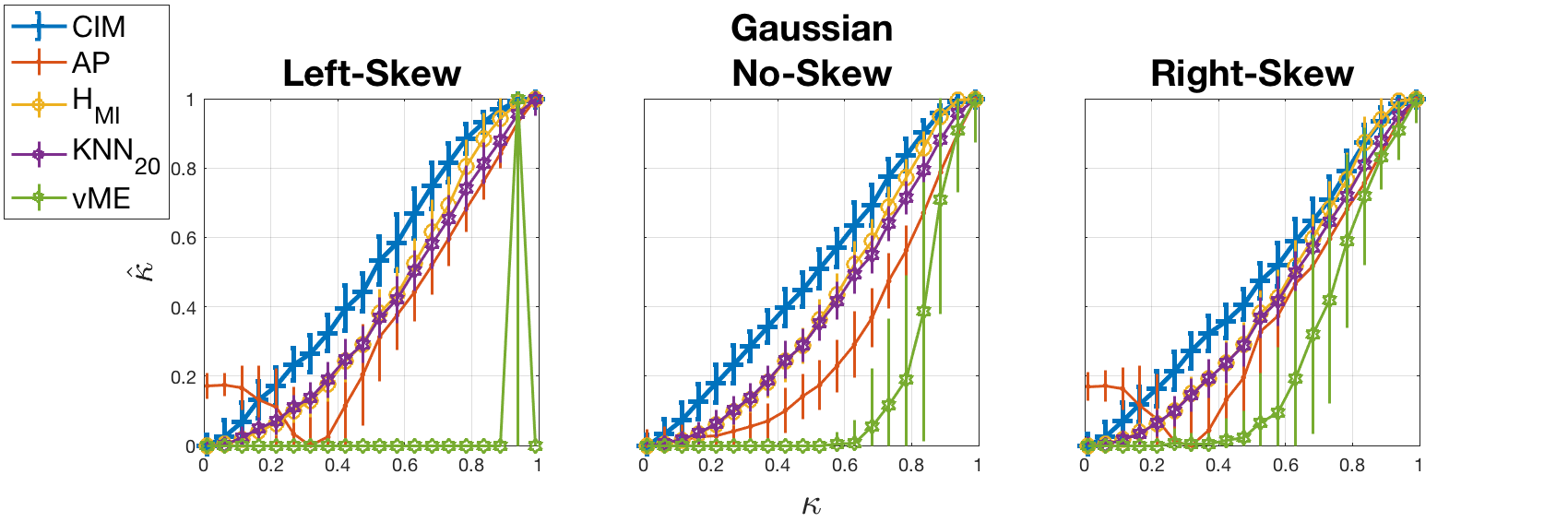}
    \end{subfigure}
    
    \begin{subfigure}[t]{1.\textwidth}
        \centering
        \includegraphics[height=1.4in]{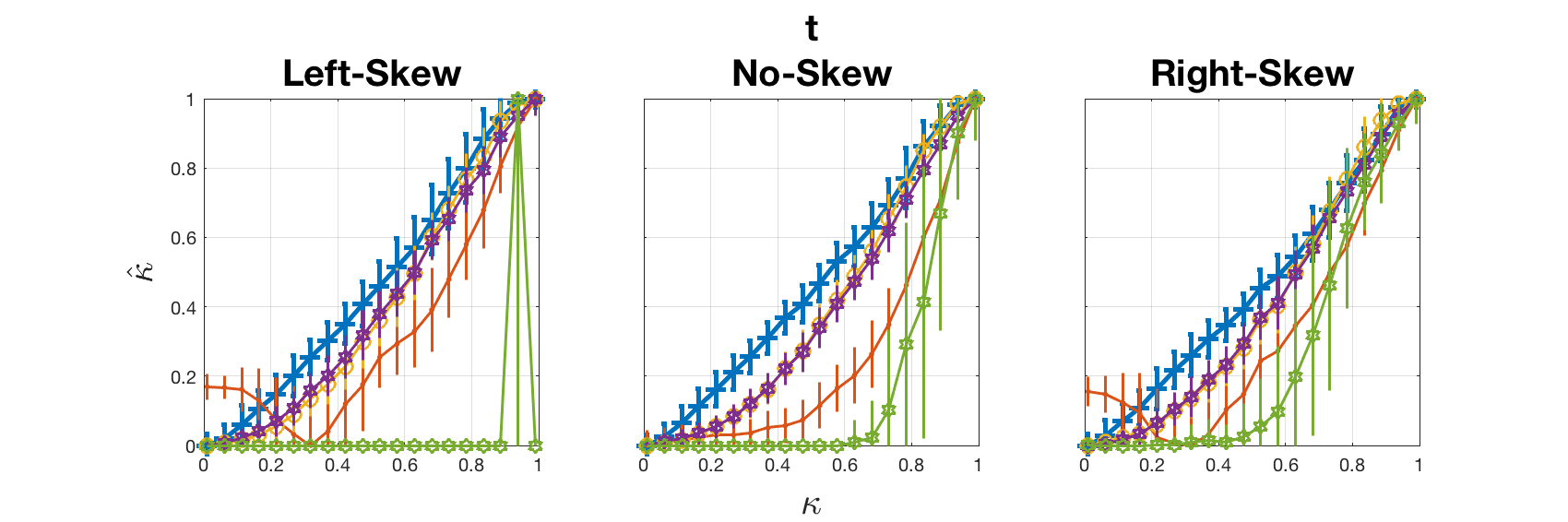}
    \end{subfigure}
    
    \begin{subfigure}[t]{1.\textwidth}
        \centering
        \includegraphics[height=1.4in]{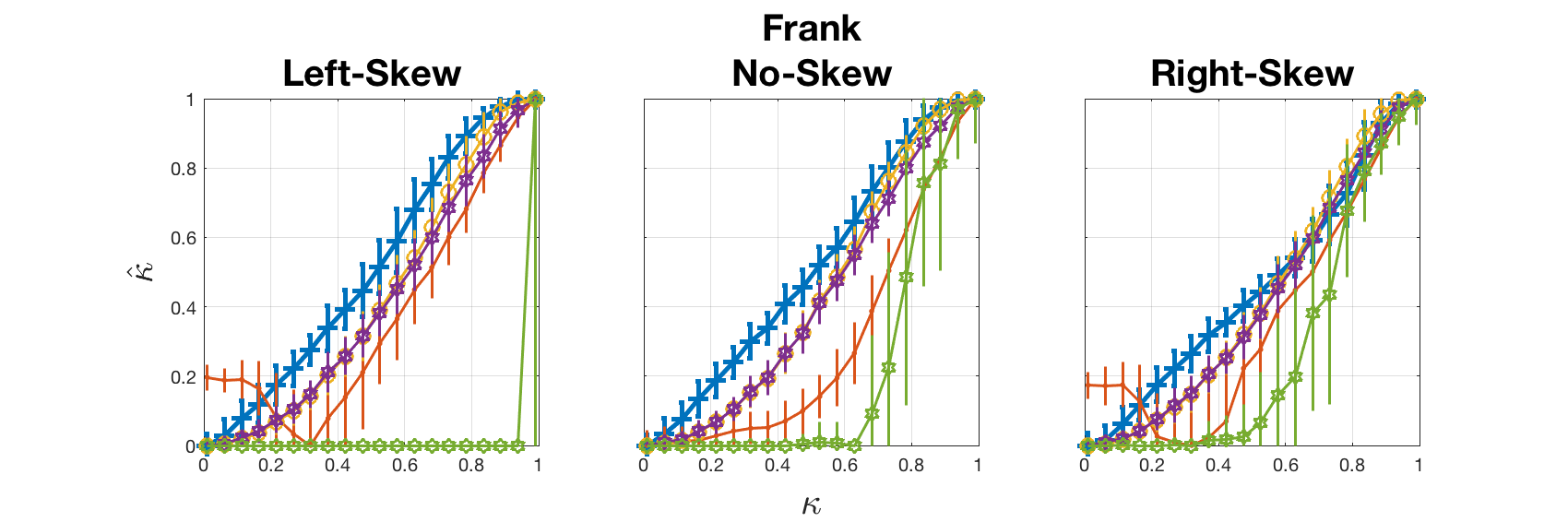}
    \end{subfigure}
    
    \begin{subfigure}[t]{1.\textwidth}
        \centering
        \includegraphics[height=1.4in]{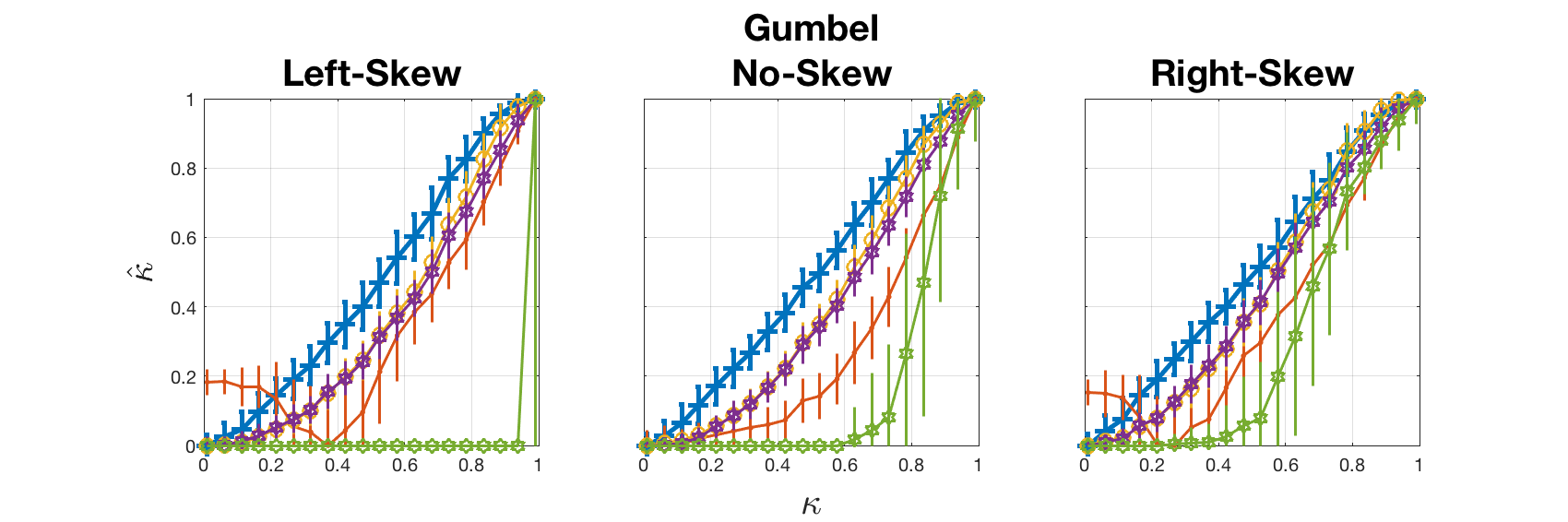}
    \end{subfigure}
    
    \begin{subfigure}[t]{1.\textwidth}
        \centering
        \includegraphics[height=1.4in]{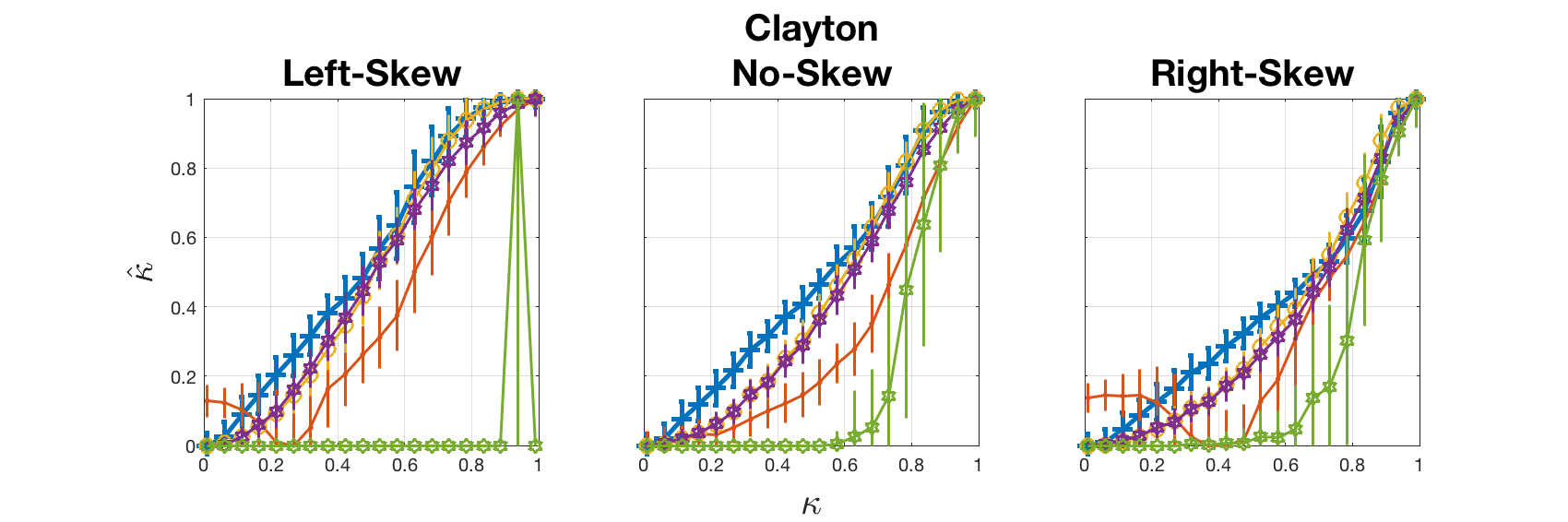}
    \end{subfigure}
    
    \caption{Response curves for \textit{kNN},\textit{vME},\textit{AP},$H_{MI}$, and \textit{CIM} for Skewed Hybrid Data.  The x-axis shows the strength of dependence, denoted by $\kappa$; here, $\kappa=0$ implies independence and $\kappa=1$ implies perfect comonotonic association.  The y-axis shows the estimated strength of dependence, denoted as $\hat{\kappa}$.  Subplots titled Left-Skew have a continuous independent variable, distributed according to the Pearson distribution with a mean of zero, standard deviation of one, and a skew of negative one.  No-Skew distributions have a continuous independent variable distributed according to a standard normal distribution, while right-skew distributions have a continuous independent variable distributed according to a Pearson distribution with a mean of zero, standard deviation of one, and a skew of positive one.  Similarly, for the positive-skew scenario, the dependent variable is a discrete random variable with a probability mass function (PMF) following the vector $[0.72,0.28]$.  The No-Skew dependent distribution is a discrete distribution following a PMF of $[0.5,0.5]$, and the negative-skew dependent distribution is a discrete distribution with PMF of $[0.28,0.72]$}
    \label{fig:synth_results}
\end{figure*}

The estimator response curves computed from the synthetic data are displayed in Fig.~\ref{fig:synth_results}.  In them, the x-axis represents the true strength of monotonic dependence, given by $\kappa$; here, $\kappa=0$ is equivalent to statistical independence, and $\kappa=1$ represents perfect comonotonic association.  The y-axis represents the estimated strength of dependence by the various estimators, denoted as $\hat{\kappa}$, with \textbf{min-max} scaling applied to the outputs of all estimators \footnote{$CIM$ is a true index of dependence, and thus always outputs a value between $0$ and $1$, where as in the theoretical sense, mutual information is unbounded on the upper end.  In our experiments, $H_{MI}$ and $KNN_{20}$ both output values between $0$ and $0.7$ across the range of tested dependencies, while $AP$ showed a maximum value of greater than $1$}.  It can be seen for all scenarios tested, $CIM$, $H_{MI}$, and $KNN_{20}$ exhibit comparable performance, with $CIM$ exhibiting slightly more linear performance across the range of compared scenarios.  Concerningly, both $vME$ and $AP$ both exhibit erratic performance in certain skewed scenarios.

\section{Application of the ERC to Mutual Information based Feature Selection}
In this section, we show how suboptimal estimation of mutual information affects feature selection and classification performance.  We begin by synthetically generating a dataset where we create $20$ independent and $20$ redundant features that explain the output class, and additionally insert $160$ random features that have no relationship to the output class.  The data is generated by first creating a $20 \times 20$ correlation matrix with linearly spaced increasing strengths of association to the output to be predicted.  Only direct correlations are modeled, so all entries are null except for the diagonal (all unity), and the last row and column, which are identical.  We test four different ranges of correlation strengths: low, medium, high, and all.  Low corresponds to correlations ranging between $0.15$ and $0.40$.  Medium corresponds to correlations ranging between $0.30$ and $0.70$.  High corresponds to correlations ranging between $0.60$ and $0.85$, and all corresponds to correlations ranging between $0.15$ and $0.85$.  These ranges of dependence strength are chosen heuristically to represent correlation strengths in real datasets.  The ranges, heuristically chosen, correspond to datasets where there is low, medium, high, and a full range of dependence strengths between the independent and explanatory variables.  These ranges were chosen to provide a good representation of what may be encountered when analyzing a real-life dataset.  In order to account for spurious correlations and noisy data collection in real-life datasets, we do not test correlation values below $0.15$ and $0.85$, respectively. 

The correlation matrices are then used to generate realizations of random variables of various marginal distributions, using Sklar's Theorem (\ref{eq:sklar1}) for both the Gaussian copula and the T copula.  We exclude Clayton, Frank, and Gumbel copulas since they cannot model individual pairwise dependence strengths between the associated features independently, even with Vine structures \cite{vinecopula}. Next, we create $20$ redundant features with linear combinations of the $20$ independent features.  Here, we restrict the combinations to multiplications and summations of pairwise independent features.  Finally, we generate $160$ random features, completing the synthetic data set.  The precise methodology to create this dataset is provided in Appendix~\ref{appendix:synth_feature_select_data_generation}.

We then apply mutual information based feature selection with the maximum relevance, minimum redundancy (MRMR) feature selection algorithm \cite{mrmr}.  Briefly, MRMR is an approach to feature selection that utilizes mutual information to assess the relevance and redundancy of a set of features, given an output prediction class.  The goal of MRMR is to solve

\begin{equation*}\label{eq:mrmr}
\argmax_{|S|=k} I(X_S,Y),
\end{equation*}
where $X_S = \left\{X_i: i \in S\right\}$, $k$ is the number of features to be selected, and $I(X,Y)$ is the mutual information between the random variables $X$ and $Y$.  However, measuring the mutual information of increasingly large dimensions of $k$ is unfeasible due to the curse of dimensionality.  MRMR attempts to overcome this by solving the optimization problem given by

\begin{equation}\label{eq:mrmr_objective}
\Phi(X_S,Y) = \frac{1}{|S|} \sum_{i \in S} I(X_i, Y) - \frac{1}{|S|^2} \sum_{i,j \in S} I(X_i,X_j).
\end{equation}
To maximize this objective, the most important feature (the feature which has the maximum mutual information with the output) is chosen first.  Then, additional features are added inductively using the function

\begin{equation}\label{eq:mrmr_solver}
\argmax_{X_j \in X \setminus S_m} I(X_j,Y) - \frac{1}{m-1} \sum_{X_i \in S_m} I(X_i,X_j).
\end{equation}
The first term in (\ref{eq:mrmr_solver}) represents the relevance of feature $X_j$ to output $Y$, and the second term represents the redundancy between the selected features $X_i$ and the current feature under consideration, $X_j$.  Because the MRMR algorithm is based on measuring mutual information between input features (continuous or discrete) and the output class (typically discrete with small cardinality), our goal in these experiments is to understand how suboptimal estimation of mutual information affects MRMR.  It is readily seen from (\ref{eq:mrmr_objective}) and (\ref{eq:mrmr_solver}) that more accurate estimation of mutual information should yield better feature selection results. 

\begin{figure*}[!t]
\centering
\includegraphics[width=.99\textwidth]{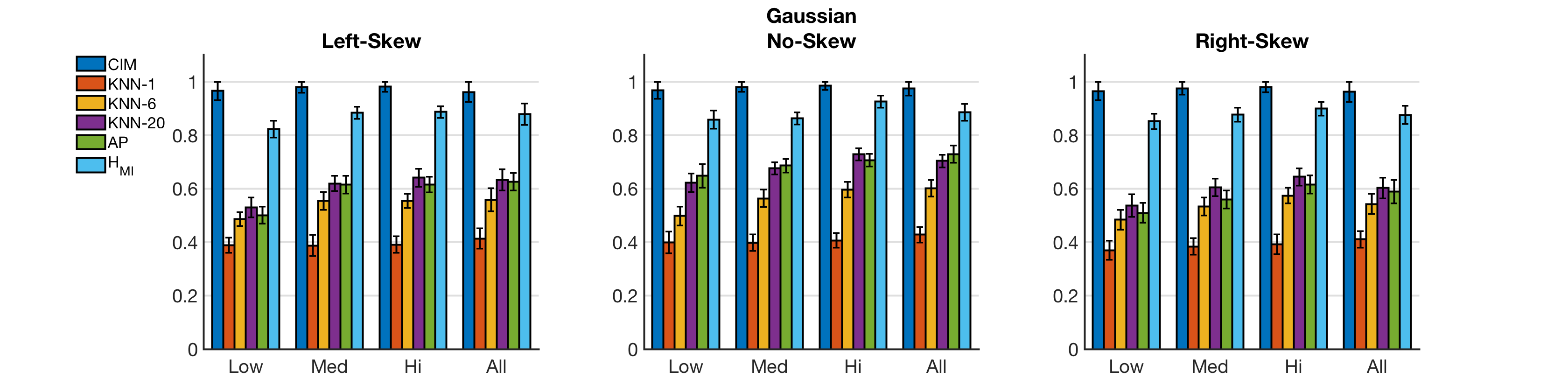}
\includegraphics[width=.99\textwidth]{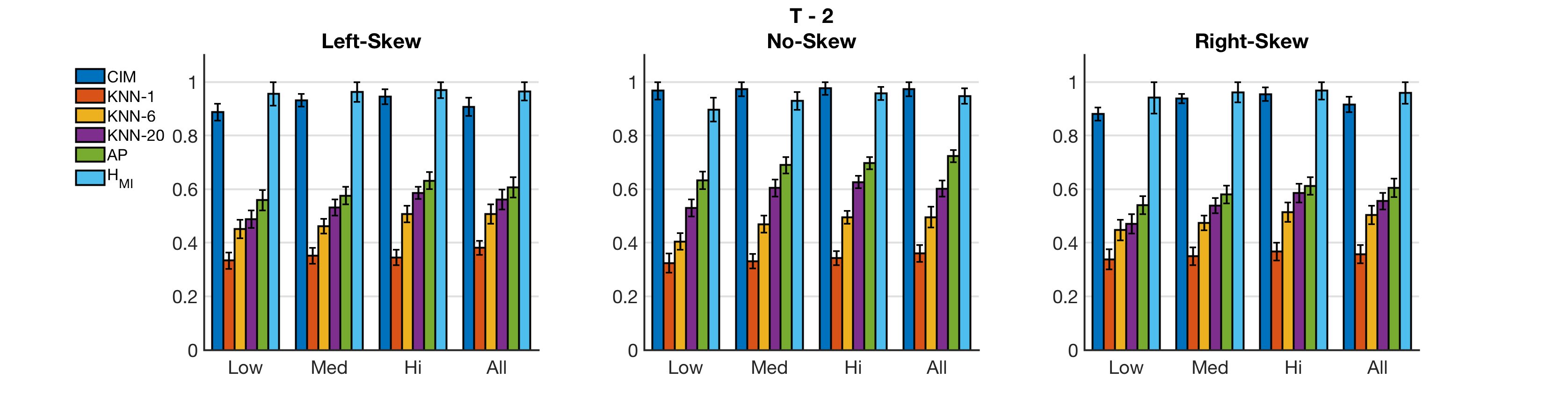}
\caption{For each group of association strength (low, medium, high, all), we count the number of correct independent features for left-skewed, no-skewed, and right-skewed data selected by MRMR for various DPI estimators, for $250$ data points, using both the Gaussian and T copulas (with 2 Degrees of Freedom).  The plots show the normalized count of the number of correct features selected.  The results show that $H_{MI}$ and $CIM$ consistently select the most relevant features.}
\label{fig:synth_feature_select}
\end{figure*}

The results of the feature selection from the synthetic datasets created are shown in Fig.~\ref{fig:synth_feature_select}.  For each group (low, medium, high, and all), we count the number of correct independent features for left-skewed, no-skewed, and right-skewed data selected by MRMR for various DPI estimators, for $250$ data points, and compute the mean and variance of the number of correct features selected over $25$ Monte-Carlo simulations.  The plots show the normalized correct number of features (normalized by the number of correct features as predicted by $CIM$).  The results show that $H_{MI}$ and $CIM$ consistently select the most relevant features.  We additionally conduct statistical preference analysis to determine whether $CIM$ is the most favorable estimator in a statistically significant way.  For each Monte-Carlo simulation, we consider the estimator which resulted in the maximum number of relevant features to be the ``prefered'' estimator; the preferences are tallied across all Monte-Carlo experiments to determine a preference table for each experiment configuration (Skew, Dependence Cluster, and Copula).  We then use the two-sided binomial test to compute the statistical significance of the preference for each estimator and experiment configuration.  The resulting p-values for $CIM$ and $H_{MI}$ are shown in Table ~\ref{tab:preference_pvals}, with the best estimator indicated in bold  The statistical preference analysis corroborates the results of the ERC computed in Fig.~\ref{fig:synth_results}.

\begin{table}
\begin{center}
\begin{tabular}{||c|c|c||}
    \hline
     \vtop{\hbox{\strut Experiment}\hbox{\strut Configuration}} & \vtop{\hbox{\strut $CIM$ p-val}\hbox{\strut Gaussian/T-2}} & 
     \vtop{\hbox{\strut $H_{MI}$ p-val}\hbox{\strut Gaussian/T-2}} \\ \hline \hline
     (Left-Skew, Low) & \textbf{0.00}/\textbf{0.00} & 0.055/0.026 \\
     (Left-Skew, Med) & \textbf{0.00}/\textbf{0.00} & 0.78/0.29 \\ 
     (Left-Skew, High) & \textbf{0.00}/\textbf{0.00} & 0.015/0.595 \\ 
     (Left-Skew, All) & \textbf{0.00}/0.055 & 1.00/\textbf{0.00} \\ 
     (No-Skew, Low) & \textbf{0.00}/\textbf{0.00} & 0.015225/0.015225 \\
     (No-Skew, Med) & \textbf{0.00}/\textbf{0.00} & 0.015225/0.015225 \\ 
     (No-Skew, High) & \textbf{0.00}/\textbf{0.00} & 0.015225/0.015225 \\ 
     (No-Skew, All) & \textbf{0.00}/\textbf{0.00} & 0.015225/0.787832 \\ 
     (Hi-Skew, Low) & \textbf{0.00}/0.000277 & 0.172126/\textbf{0.000054} \\
     (Hi-Skew, Med) & \textbf{0.00}/\textbf{0.000009} & 0.026190/0.001234 \\ 
     (Hi-Skew, High) & \textbf{0.00}/\textbf{0.00} & 0.416732/0.787832 \\ 
     (Hi-Skew, All) & \textbf{0.00}/0.001234 & 0.055215/\textbf{0.000009} \\ 
     \hline \hline
\end{tabular}
\end{center}
\caption{Statistical preference analysis for synthetic feature selection results.  The tables show p-values from a two-sided binomial test for the $CIM$ and $H_{MI}$ estimators, where each estimator for each experiment configuration is ``preferred'' if it selected the most number of relevant features.  We see that $CIM$ performs favorably compared to $H_{MI}$, and we can infer from Fig.~\ref{fig:synth_feature_select} that this extends to the other estimators tested.}
\label{tab:preference_pvals}
\end{table}

Next, we check the empirical transferrability of the results in Fig.~\ref{fig:synth_feature_select} to real world datasets.  To accomplish this, we take seven real world datasets, four originating from the NIPS 2003 feature selection challenge which which have binary response variables \cite{nips_datasets}, and three from additional sources which response variables with a support greater than two.  For each dataset, apply the MRMR algorithm to select the most relevant features.  Unlike the synthetic dataset generated above, we do not have knowledge of which features predict the output and which ones are randomly inserted.  Thus, we test our hypothesis of accurate feature selection by applying classification algorithms on the selected features; our assumption is that higher classification performance implies a better estimator of mutual information in the context of rank ordering dependencies because the same classification and feature selection algorithms are used across all tests.  The estimators compared are \textit{kNN-1}, \textit{kNN-6}, \textit{kNN-20}, \textit{vME}, \textit{AP}, \textit{CIM}, and $H_{MI}$; these are chosen because they are proven to satisfy the DPI assumption required by MRMR.  Using the selected features for each estimator, we then apply three classification algorithms (KNN, SVM, and Random Forest) and score the classification performance using only the selected features on a validation dataset.For the NIPS 2003 datasets, this process is repeated when different amounts of data from the positive class are dropped, creating skewed output class distributions. The deliberate skewing of the output class allows us to understand how well the ERCs computed in Fig.~\ref{fig:synth_results} translate to real-world skewed dependent variable scenarios.  We do not apply the skewing to the three datasets which multiclass classification, as the number of combinations of which label to skew grows exponentially.

The NIPS 2003 datasets, Arcene, Dexter, Madelon, and Gisette, are binary classification datasets, where the input features are continuous variables, and the output is a discrete variable that can take on two values.  These datasets were chosen because from the perspective of rank ordering the strength of association between random variables, this presents the most ``difficult'' case.  From a practicality perspective, this case is also highly relevant to machine learning problems, where predictive features are often continuous, but the output class to be predicted has only a small number of unique outcomes.  Additionally, they represent datasets of various sample sizes and are from different fields in science.  The Arcene dataset is $100$ samples and contains $10000$ features; $7000$ of the features are data collected from the mass-spectrometers, and $3000$ features are probes, having no predictive power in predicting whether the mass spectrometric data originated  from cancerous or non-cancerous cells.  The Dexter dataset has $300$ samples, and contains $20000$ features; $9947$ of the features are bag of words features extracted from Reuters news articles, and $10053$ are probes, having no predictive power of determining whether news articles are about ``corporate acquisitions''.   The Madelon dataset has $2000$ samples and is an artificial dataset with $500$;  $20$ highly nonlinear features contain information to identify which points on a five-dimensional hypercube the points belong to, while $480$ features are probes and have no predictive power.  Finally, the Gisette dataset is $3000$ samples of $5000$ features; $2500$ features represent random pixels sampled from images of the digits $4$ and $9$ and contain the predictive power to distinguish between the two, while $2500$ features are probes and contain no predictive power for this task.

\begin{figure*}
    \centering
    \begin{subfigure}[t]{1\textwidth}
        \centering
        \includegraphics[height=1.5in]{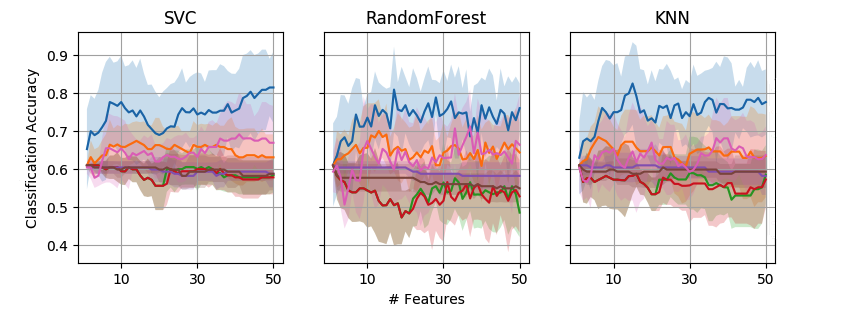}
        \caption{Arcene - Skew=$0.5$}
        \label{fig:arcene_0_5}
    \end{subfigure}
    
    \begin{subfigure}[t]{1\textwidth}
        \centering
        \includegraphics[height=1.5in]{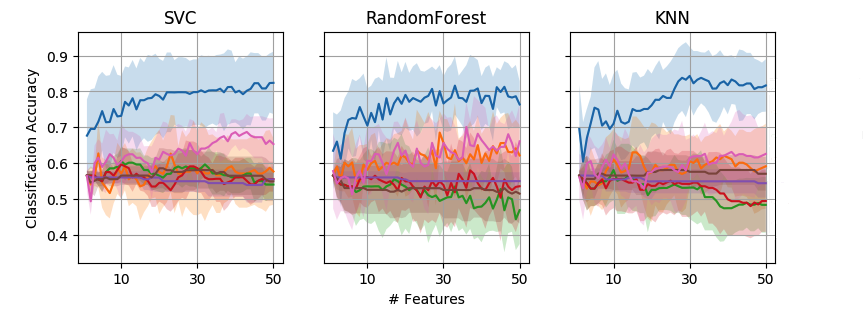}
        \caption{Arcene - Skew=$0.75$}
        \label{fig:arcene_0_75}
    \end{subfigure}
    
    \begin{subfigure}[t]{1\textwidth}
        \centering
        \includegraphics[height=1.5in]{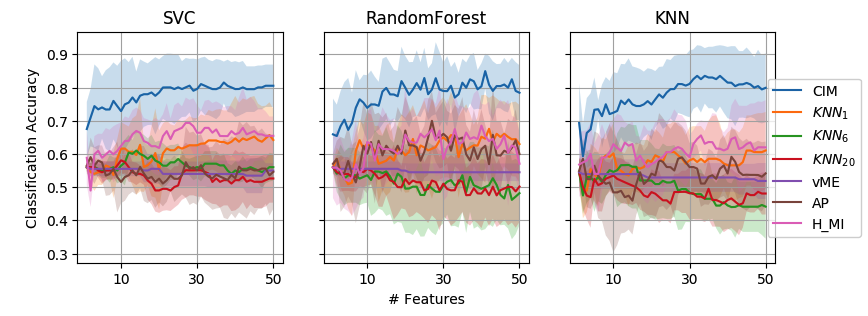}
        \caption{Arcene - No Skew}
        \label{fig:arcene_full}
    \end{subfigure}
    
    \caption{Results for feature selection on the Arcene dataset.  The results of feature selection and subsequent classification by \textit{SVC}, \textit{RandomForest}, and \textit{kNN} classifiers, for the output class balances indicated in the caption.  The skew amount indicates the fraction of labels that are $1$, compared to the the number of labels which are $0$.}
    \label{fig:real_world_results_arcene}
\end{figure*}

\begin{figure*}
    \centering
    \begin{subfigure}[t]{1\textwidth}
        \centering
        \includegraphics[height=1.5in]{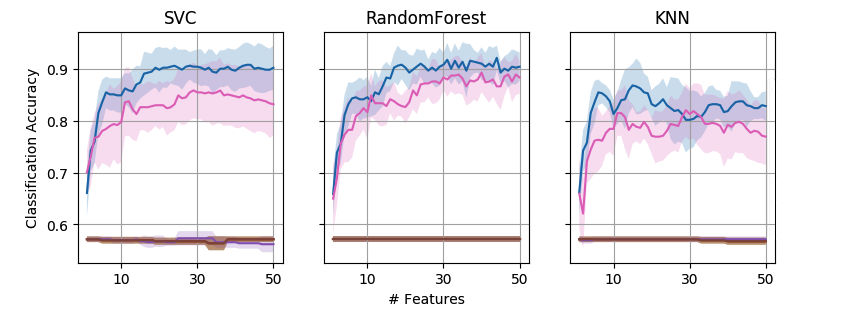}
        \caption{Dexter - Skew=$0.5$}
        \label{fig:dexter_0_5}
    \end{subfigure}
    
    \begin{subfigure}[t]{1\textwidth}
        \centering
        \includegraphics[height=1.5in]{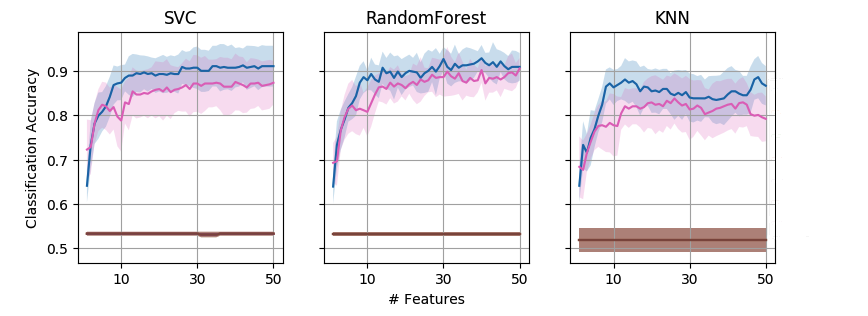}
        \caption{Dexter - Skew=$0.75$}
        \label{fig:dexter_0_75}
    \end{subfigure}
    
    \begin{subfigure}[t]{1\textwidth}
        \centering
        \includegraphics[height=1.5in]{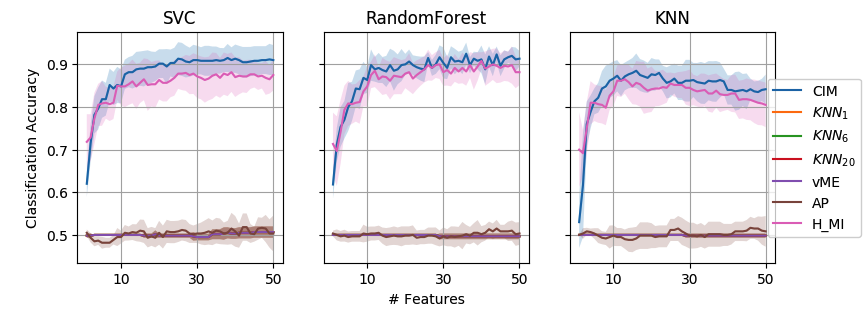}
        \caption{Dexter - No Skew}
        \label{fig:dexter_full}
    \end{subfigure}
    
    \caption{Results for feature selection on the Dexter dataset.  The results of feature selection and subsequent classification by \textit{SVC}, \textit{RandomForest}, and \textit{kNN} classifiers, for the output class balances indicated in the caption.  The skew amount indicates the fraction of labels that are $1$, compared to the the number of labels which are $0$.}
    \label{fig:real_world_results_dexter}
\end{figure*}

\begin{figure*}
    \centering
    \begin{subfigure}[t]{1\textwidth}
        \centering
        \includegraphics[height=1.5in]{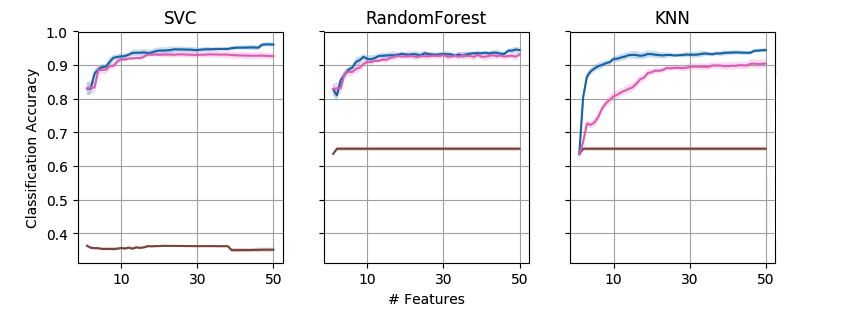}
        \caption{Gisette - Skew=$0.5$}
        \label{fig:gisette_0_5}
    \end{subfigure}
    
    \begin{subfigure}[t]{1\textwidth}
        \centering
        \includegraphics[height=1.5in]{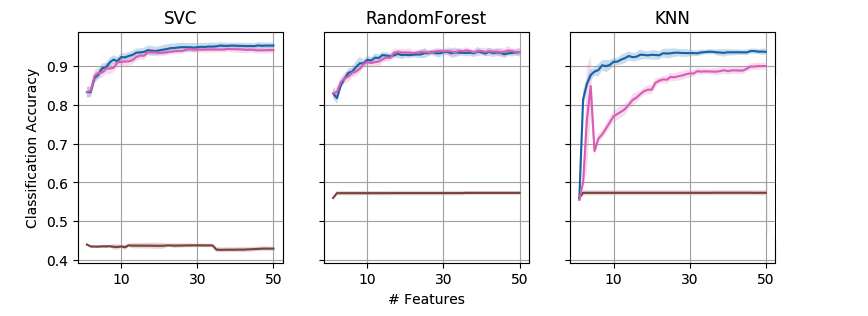}
        \caption{Gisette - Skew=$0.75$}
        \label{fig:gisette_0_75}
    \end{subfigure}
    
    \begin{subfigure}[t]{1\textwidth}
        \centering
        \includegraphics[height=1.5in]{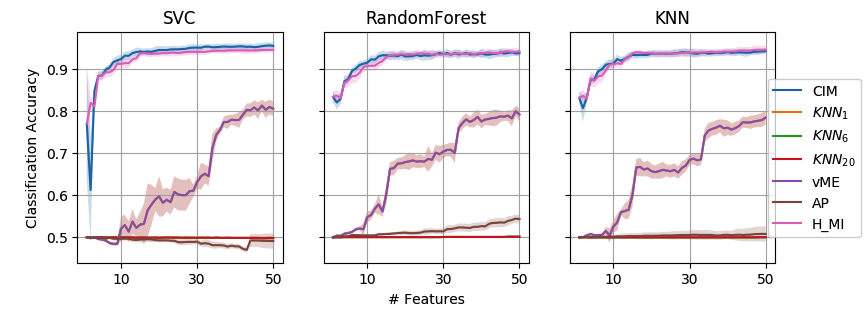}
        \caption{Gisette - No Skew}
        \label{fig:gisette_full}
    \end{subfigure}
    
    \caption{Results for feature selection on the Gisette dataset.  The results of feature selection and subsequent classification by \textit{SVC}, \textit{RandomForest}, and \textit{kNN} classifiers, for the output class balances indicated in the caption.  The skew amount indicates the fraction of labels that are $1$, compared to the the number of labels which are $0$.}
    \label{fig:real_world_results_gisette}
\end{figure*}

\begin{figure*}
    \centering
    \begin{subfigure}[t]{1\textwidth}
        \centering
        \includegraphics[height=1.5in]{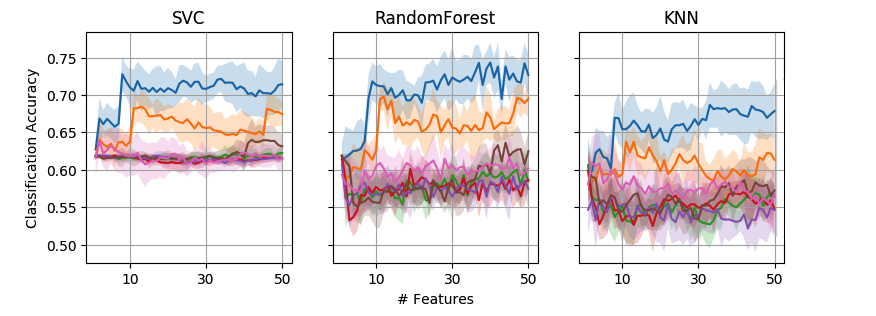}
        \caption{Madelon - Skew=$0.5$}
        \label{fig:madelon_0_5}
    \end{subfigure}
    
    \begin{subfigure}[t]{1\textwidth}
        \centering
        \includegraphics[height=1.5in]{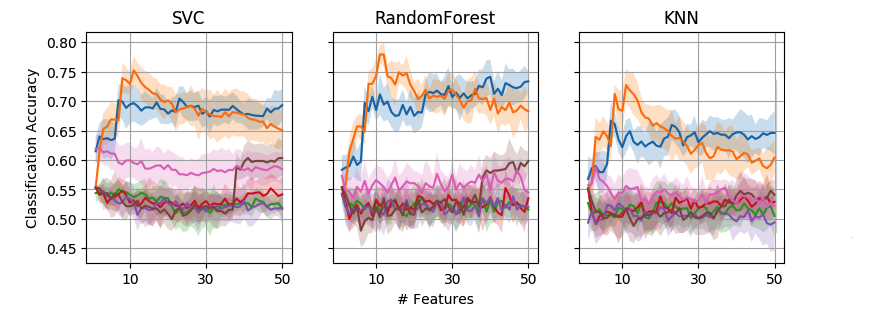}
        \caption{Madelon - Skew=$0.75$}
        \label{fig:madelon_0_75}
    \end{subfigure}
    
    \begin{subfigure}[t]{1\textwidth}
        \centering
        \includegraphics[height=1.5in]{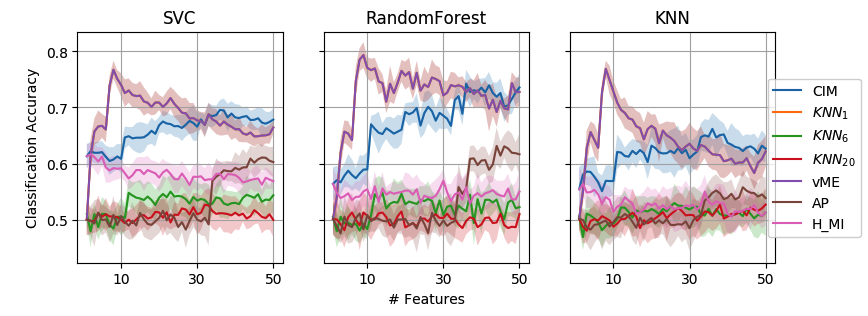}
        \caption{Madelon - No Skew}
        \label{fig:madelon_full}
    \end{subfigure}
    
    \caption{Results for feature selection on the Madelon dataset.  The results of feature selection and subsequent classification by \textit{SVC}, \textit{RandomForest}, and \textit{kNN} classifiers, for the output class balances indicated in the caption.  The skew amount indicates the fraction of labels that are $1$, compared to the the number of labels which are $0$.}
    \label{fig:real_world_results_madelon}
\end{figure*}

The results for our experiments conducted with the NIPS 2003 dataset are displayed in Figs.~\ref{fig:real_world_results_arcene},~\ref{fig:real_world_results_dexter},~\ref{fig:real_world_results_gisette},~\ref{fig:real_world_results_madelon}.  For each dataset, we show the 10-fold cross validation score of \textit{kNN}, \textit{SVC}, and \textit{Random Forest} classifiers as we increase the number of features that were selected, in order of importance as provided by the MRMR algorithm for each DPI satisfying estimator.  We show the results for each dataset, where we skew the number of positive examples to be $50\%$, $75\%$, and $100\%$ of the number of negative examples.  The output class distribution for each simulation is indicated by the respective caption of each subplot.  It is seen that for all datasets tested, the \textit{CIM} estimator compares atleast as favorably as all other estimators of mutual information considered.  The results corroborate the findings in Section~\ref{sec:synthetic_simulations}, where it was seen that with synthetic test vectors, the \textit{CIM} estimator compares favorably to other DPI satisfying measures of the strength of association for hybrid random variables.

Finally, we test the same estimators against three real world datasets where the support of the output class is greater than 2; i.e. corresponding to the multiclass classification scenario.  The datasets chosen were arrhythmia, colonoscopy, and a cancer rna sequencing dataset.  The arrhythmia dataset consists of $452$ samples and contains $279$ features describing patient statistics and a representation of the electrocardiogram (ECG) measurements; the task is to identify samples into one of sixteen classes of cardiac arrhythmia.  The colonoscopy dataset consists of $76$ samples and contains $698$ features representing information regarding gastrointestinal legions; the task is to identify samples as belonging to one of three types of lesions: hyperplasic, adenoma, and serrated adenoma.  The cancer ran sequencing dataset consists of $801$ samples and contains $20351$ features representing random extractions of gene expressions of patients having different types of tumors; the task is to identify which one of the five types of tumors each of these gene expressions corresponds to.

\begin{figure*}
    \centering
    \begin{subfigure}[t]{1\textwidth}
        \centering
        \includegraphics[height=1.5in]{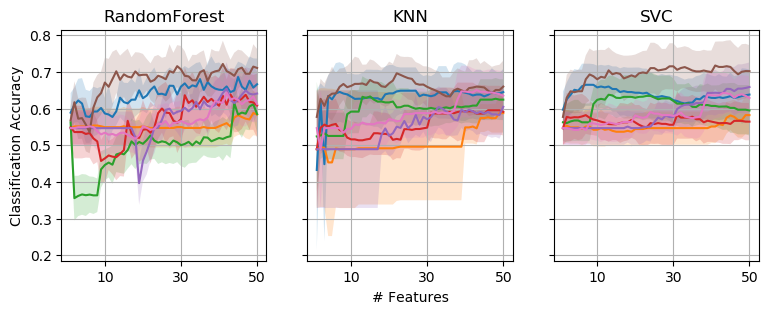}
        \caption{Arrhythmia}
        \label{fig:arrhythmia}
    \end{subfigure}
    
    \begin{subfigure}[t]{1\textwidth}
        \centering
        \includegraphics[height=1.5in]{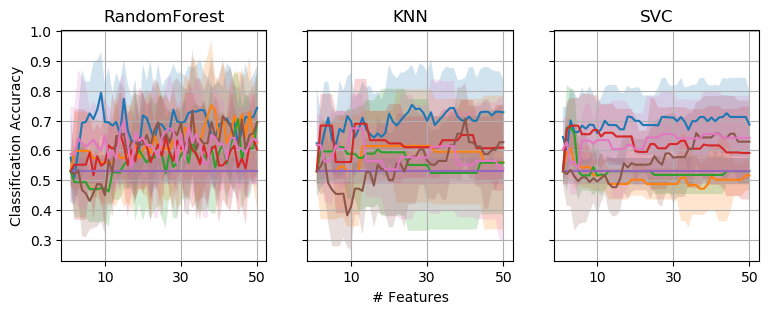}
        \caption{Colonoscopy}
        \label{fig:colon}
    \end{subfigure}
    
    \begin{subfigure}[t]{1\textwidth}
        \centering
        \includegraphics[height=1.5in]{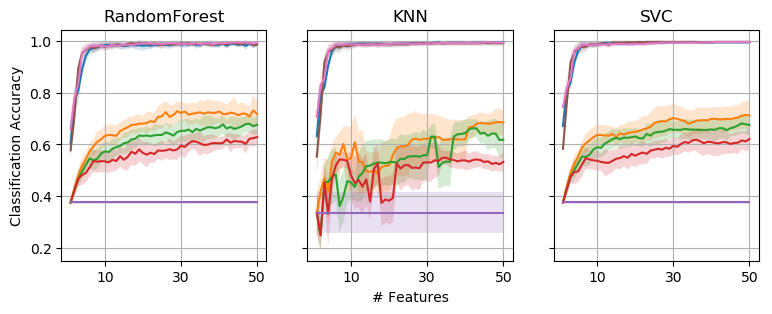}
        \caption{Cancer RNA Seq}
        \label{fig:cancer_rna_seq}
    \end{subfigure}
    
    \caption{Real World Data results for Feature Selection.  For each dataset, we show the results of feature selection and subsequent classification by \textit{SVC}, \textit{RandomForest}, and \textit{kNN} classifiers.}
    \label{fig:real_world_results2}
\end{figure*}

The results for our experiments conducted with the multiclass classification datasets are displayed in Fig.~\ref{fig:real_world_results2}.  For each dataset, we show the 10-fold cross validation score of \textit{kNN}, \textit{SVC}, and \textit{Random Forest} classifiers as we increase the number of features that were selected, in order of importance as provided by the MRMR algorithm for each DPI satisfying estimator.  It is seen that for all datasets tested, the \textit{CIM} estimator compares atleast as favorably as all other estimators of mutual information considered.  The results, again, corroborate the findings in Section~\ref{sec:synthetic_simulations}.

Finally, we conduct statistical preference analysis on the real-world datasets.  Here, we consider an estimator to be preferred if it produces the maximum classification accuracy when using the top fifty features selected.  From Figs.~\ref{fig:real_world_results_arcene},~\ref{fig:real_world_results_dexter},~\ref{fig:real_world_results_gisette},~\ref{fig:real_world_results_madelon}, and ~\ref{fig:real_world_results2}, we see that $CIM$ satisfies this criterion thirty out of the thirty-nine experiments conducted.  According to the two-sided binomial test, this corresponds to a p-value of 0.00, indicating that $CIM$ compares favorably to the other estimators of DPI compared, in a statistically significant manner.

\section{Conclusions}
In this paper, we have introduced a new concept for evaluating the performance of an estimator of the strength of association between random variables, the estimator response curve.  We show that existing empirical properties for measuring the performance of an estimator are inadequate, and explain how the estimator response curve fills this gap.  We then explain a copula based methodology for measuring the response curve of an estimator, and apply this methodology to estimate the response curves of various estimators of the strength of association which satisfy the DPI criterion.  Comparing the estimator response curves, we see that the \textit{CIM} estimator performs best across the board in the hybrid random variable scenario, where data may be skewed.  We then test these various estimators with real world data.  The simulations show that the estimator response curves are a good indicator of how an estimator may perform in a scenario where the strengths of associations need to be ranked, as in feature selection and classification.


%
\pagebreak

\appendices
\section{Metholodogy for creating Synthetic Feature Selection Dataset}\label{appendix:synth_feature_select_data_generation}
The following algorithm details the procedure we take in order to generate the synthetic feature selection dataset.

\begin{algorithm}[H]
 \caption{Algorithm for generating synthetic feature selection dataset}
 \begin{algorithmic}[1]
 \REQUIRE $corr$
 \ENSURE  $X$
  \IF {$corr$==``low''}
    \STATE $\boldsymbol{\rho} \gets $ \underline{\textbf{\textit{linspace}}}(0.15,0.40,20)
  \ELSIF{$corr$==``med''}
    \STATE $\boldsymbol{\rho} \gets $ \underline{\textbf{\textit{linspace}}}(0.30,0.70,20)
  \ELSIF{$corr$==``hi''}
    \STATE $\boldsymbol{\rho} \gets $ \underline{\textbf{\textit{linspace}}}(0.60,0.85,20)
  \ELSE
    \STATE $\boldsymbol{\rho} \gets $ \underline{\textbf{\textit{linspace}}}(0.15,0.85,20)
  \ENDIF
 \STATE $R \gets \mathbb{1}^{20+1 \times 20+1}$
 \STATE $R[21,:] \gets \boldsymbol{\rho}$
 \STATE $R[:,21] \gets \boldsymbol{\rho}$
 \STATE $U,V \gets $ \underline{\textbf{\textit{copularnd}}}($R$)
 \RETURN $X$ 
 \end{algorithmic} 
 \end{algorithm}

\pagebreak

\section*{Acknowledgments}
The authors would like to thank the Hume Center at Virginia Tech for its support.  We additionally thank the anonymous reviewers who have provided valuable feedback to help improve the quality of this manuscript.

\clearpage



%

\bibliographystyle{elsarticle-num-names}
\bibliography{cim_references}

\end{document}